\documentclass[runningheads]{llncs}
\usepackage[T1]{fontenc}
\usepackage{graphicx}
\usepackage{adjustbox} 
\usepackage{enumitem}
\usepackage{multirow}
\usepackage{times}
\usepackage{graphicx}
\usepackage{float}
\usepackage{subfigure}
\usepackage[misc]{ifsym}
\usepackage{lmodern}
\usepackage{anyfontsize}

\begin{document}
\title{HW-TSC's Submission to the CCMT 2024 Machine Translation Tasks}
%
%
\author{Zhanglin Wu\inst{1}\orcidID{0000-0002-2920-0773},Yuanchang Luo, Daimeng Wei, Jiawei Zheng, Bin Wei, Zongyao Li, Hengchao Shang, Jiaxin Guo, Shaojun Li, Weidong Zhang, Ning Xie, Hao Yang}
%
\authorrunning{Zhanglin et al.}

\institute{Huawei Translation Service Center, Beijing, China\\
\email{\{wuzhanglin2,luoyuanchang1,weidaimeng,zhengjiawei15,\\
weibin29,lizongyao,shanghengchao,guojiaxin1,lishaojun18,\\
zhangweidong17,nicolas.xie,yanghao30\}@huawei.com}}

%
\maketitle              
\begin{abstract}
This paper presents the submission of Huawei Translation Services Center (HW-TSC) to machine translation tasks of the 20th China Conference on Machine Translation (CCMT 2024). We participate in the bilingual machine translation task and multi-domain machine translation task. For these two translation tasks, we use training strategies such as regularized dropout, bidirectional training, data diversification, forward translation, back translation, alternated training, curriculum learning, and transductive ensemble learning to train neural machine translation (NMT) models based on the deep Transformer-big architecture. Furthermore, to explore whether large language model (LLM) can effectively improve the translation quality of NMT models, we use supervised fine-tuning (SFT) to train llama2-13b as an Automatic post-editing (APE) model to improve the translation results of the NMT model on the multi-domain machine translation task. By using these plyometric strategies, our submission achieves a competitive result in the final evaluation.
\end{abstract}

\section{Introduction}\label{sec1}

Neural machine translation (NMT) \cite{bib1,bib2,bib3} allows translation systems to be trained end-to-end without dealing with issues like word alignment, translation rules, and complex decoding algorithms that characterize statistical machine translation (SMT) \cite{bib4} systems. Although NMT develops rapidly in recent years, it relies heavily on big data - large-scale, high-quality bilingual corpora. Due to the cost and scarcity of real corpora, synthetic data plays an important role in improving translation quality. Existing methods for synthesizing data in NMT focus on leveraging monolingual data during training. Among them, data diversification \cite{bib5}, forward translation \cite{bib6}, and back translation \cite{bib7} are widely used to generate synthetic bilingual corpora. Such synthetic data can be used to improve the performance of NMT models. Although synthetic data is efficient, it inevitably contains noise and erroneous translations. Alternated training \cite{bib8} introduces real data as a guide and alternately uses synthetic data and real data during the training process, which can prevent the training of the NMT model from being interfered with by noisy synthetic data. Another direction to improve the performance of NMT models is to use more efficient training strategies. Methods such as regularized dropout \cite{bib9}, bidirectional training \cite{bib10}, and curriculum learning \cite{bib11} allow the NMT model to more effectively utilize limited data during the training process, while transductive ensemble learning \cite{bib12} can aggregate the translation capabilities of multiple models into one model.

The powerful capabilities of LLM in logical reasoning and language generation promote the further development of machine translation. Despite the superior performance of translation models, existing models usually use beam search decoding \cite{bib13} and top-1 hypothesis selection for inference. These techniques struggle to fully exploit the rich information in various N-best hypotheses, making them suboptimal for translation tasks that require a single high-quality output sequence. GenTranslate \cite{bib14} utilizes the rich language knowledge and powerful reasoning capabilities of large language model (LLM) \cite{bib15,bib16,bib17,bib18} to integrate the rich information in the N-best list from the basic model, generating higher-quality translation results.

To promote academic exchanges and connections between domestic and foreign research units and relevant industry partners, and to jointly advance the development of machine translation research and technology, we participate in the bilingual machine translation tasks and multi-domain machine translation tasks organized by CCMT2024. For these two translation tasks, we use training strategies such as regularized dropout \cite{bib9}, bidirectional training \cite{bib10}, data diversification \cite{bib5}, forward translation \cite{bib6}, back translation \cite{bib7}, alternated training \cite{bib8}, curriculum learning \cite{bib11}, and transductive ensemble learning \cite{bib12} to train neural machine translation (NMT) models based on the deep Transformer-big architecture \cite{bib19,bib20,bib21,bib22}. Additionally, drawing inspiration from the GenTranslate method \cite{bib14}, we utilize supervised fine-tuning (SFT) \cite{bib23} to train llama2-13b\footnote{\url{https://huggingface.co/meta-llama/Llama-2-13b-chat-hf}} as an automatic post-editing (APE) model \cite{bib24}, aimed at enhancing the translation outputs of NMT models in the multi-domain machine translation task.

This paper expands on the details of our translation system in different translation tasks. The structure of the remaining sections is as follows: Section 2 describes the data size and data pre-processing; Section 3 provides an overview of our NMT system; Section 4 explains our APE system; Section 5 presents the parameter settings and experimental results; Section 6 summarizes our systems.

\section{Dataset}\label{sec2}
    
\subsection{Data Size}

According to the requirements of the CCMT 2024 outline, we train the NMT system from scratch on the bilingual translation task using the officially provided data. Table \ref{data1} shows the training data size for each language pair of the bilingual MT task after data pre-processing. These language pairs include English$\rightarrow$Chinese (en$\rightarrow$zh), Chinese$\rightarrow$English (zh$\rightarrow$en), Mongolian$\rightarrow$Chinese ( mn$\rightarrow$zh), Tibetan$\rightarrow$Chinese (ti$\rightarrow$zh), and Uyghur$\rightarrow$Chinese (uy$\rightarrow$zh). It should be noted that in the en$\rightarrow$zh and zh$\rightarrow$en translation tasks, since the training data of WMT 2024 is shared with CCMT 2024, we additionally use the training data provided by the WMT 2024 general MT task to scale up the training data.

\begin{table}[ht]
\begin{center}
\caption{\centering Data size for each bilingual MT task after data pre-processing}\label{data1}
\begin{tabular}{|lccc|c|c|c|c|}
\hline
 &&& en$\rightarrow$zh & zh$\rightarrow$en & mn$\rightarrow$zh & ti$\rightarrow$zh & uy$\rightarrow$zh \\
\hline
Bilingual &&& 25.12M & 25.12M & 1.24M & 0.97M & 0.16M\\
Source Monolingual &&& 50M & 50M  & - & - & -\\
Target Monolingual &&& 50M & 50M  & 4.89M & 4.89M & 4.89M\\
\hline
\end{tabular}
\end{center}
\end{table}

Table \ref{data2} shows the training data size for the multi-domain machine translation task after data preprocessing. For this zh$\rightarrow$en task, the official does not provide any training data, but allows participating units to construct their own training data. Therefore, we use the training data from the bilingual translation task and also collect 33.36 million high-quality domain-related bilingual data.

\begin{table}[ht]
\begin{center}
\caption{\centering Data size for multi-domain MT task after data pre-processing}\label{data2}
\begin{tabular}{|l|ccc|cc|cc|}
\hline
 &&& Bilingual && Source Monolingual && Target Monolingual \\
\hline
zh$\rightarrow$en &&& 58.48M && 50M  && 50M  \\
\hline
\end{tabular}
\end{center}
\end{table}

\subsection{Data Pre-processing}

The data pre-processing process is as follows:

\begin{itemize}[itemindent=2pt]
    \item Remove duplicate sentences or sentence pairs.
    \item Remove invisible characters and xml escape characters.
    \item Convert full-width symbols to half-width symbols.
    \item Use jieba\footnote{\url{https://github.com/fxsjy/jieba}} to pre-segment Chinese sentences.
    \item Use mosesdecoder\footnote{\url{https://github.com/moses-smt/mosesdecoder}} to normalize English punctuation.
    \item Use opencc\footnote{\url{https://github.com/BYVoid/OpenCC}} to convert traditional Chinese to simplified Chinese.
    \item Use fasttext\footnote{\url{https://github.com/facebookresearch/fastText}} to filter other language sentences.
    \item Use fast\_align\footnote{\url{https://github.com/clab/fast\_align}} to filter poorly aligned sentence pairs.
    \item Filter out sentences with more than 150 tokens in bilingual data.
    \item Split long sentences in monolingual data into multiple short sentences.
    \item Filter out sentence pairs with token ratio greater than 4 or less than 0.25.
    \item When performing subword segmentation, joint Byte Pair Encoding\footnote{\url{https://github.com/soaxelbrooke/python-bpe}} \cite{bib25} is used for mn$\rightarrow$zh, ti$\rightarrow$zh and uy$\rightarrow$zh translation tasks, and joint sentencepiece\footnote{\url{https://github.com/google/sentencepiece}} \cite{bib26} is used for zh$\rightarrow$en and en$\rightarrow$zh translation tasks.
\end{itemize}

\section{NMT System}\label{sec3}

\subsection{System Overview}

Transformer is the state-of-the-art model structure in recent MT evaluations. There are two parts of research to improve this kind: the first part uses wide networks (eg: Transformer-Big \cite{bib27}), and the other part uses deeper language representations (eg: Deep Transformer \cite{bib28}). For all MT tasks, we combine these two improvements, adopting the Deep Transformer-Big \cite{bib19,bib20,bib21,bib22} model structure to train the NMT system. Deep Transformer-Big uses pre-layer normalization, features 25-layer encoder, 6-layer decoder, 16-heads self-attention, 1024-dimensional word embedding and 4096-dimensional ffn embedding.

\begin{figure}[H] 
\centering
\includegraphics[width=80mm]{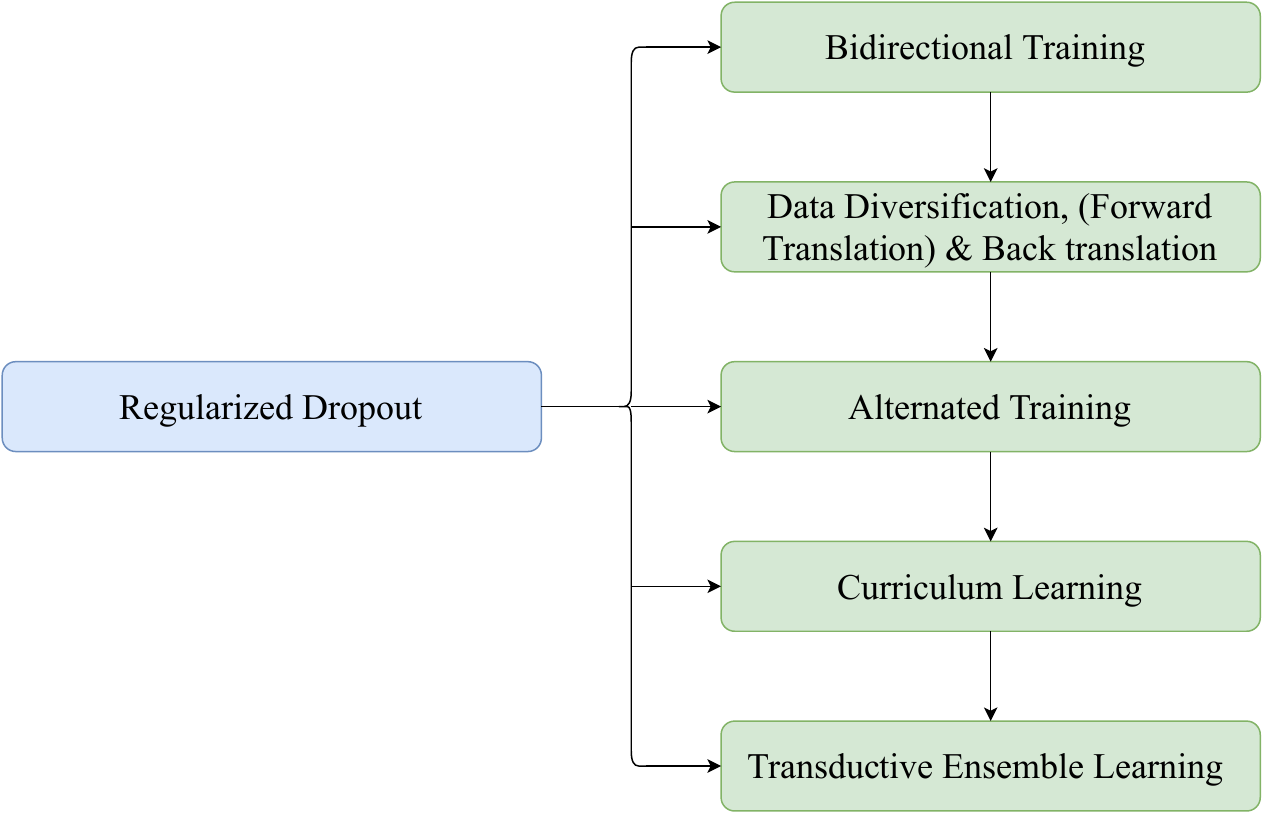}
\caption{\centering The overall training flow chart of our NMT system.}
\label{Bilingual_Training}
\end{figure}

Fig. \ref{Bilingual_Training} shows the overall training flow chart of our NMT system on the bilingual machine translation task and multi-domain machine translation task. We use training strategies such as regularized dropout \cite{bib9}, bidirectional training \cite{bib10}, data diversification \cite{bib5}, forward translation \cite{bib6}, back translation \cite{bib7}, alternated training \cite{bib8}, curriculum learning \cite{bib11}, and transductive ensemble learning \cite{bib12} to train NMT models based on the deep Transformer-big architecture \cite{bib19,bib20,bib21}. Since forward translation relies on source monolingual and mn$\rightarrow$zh, ti$\rightarrow$zh and uy$\rightarrow$zh translation tasks do not provide source monolingual, we do not use forward translation on these three tasks. Furthermore, our choice of back translation methods varies across different tasks. For en$\rightarrow$zh and zh$\rightarrow$en tasks where forward translation is available, we use sampling back translation (ST) \cite{bib29}, and for other tasks we use tagged back translation (Tagged BT) \cite{bib30}.

\subsection{Regularized Dropout}

Dropout \cite{bib31} is a widely used technique for regularizing deep neural network training, which is crucial to prevent over-fitting and improve the generalization ability of deep models. Dropout performs implicit ensemble by simply dropping a certain proportion of hidden units from the neural network during training, which may cause an unnegligible inconsistency between training and inference. Regularized Dropout\footnote{\url{https://github.com/dropreg/R-Drop}} (R-Drop) \cite{bib9} is a simple yet more effective alternative to regularize the training inconsistency induced by dropout. Concretely, in each mini-batch training, each data sample goes through the forward pass twice, and each pass is processed by a different sub model by randomly dropping out some hidden units. R-Drop forces the two distributions for the same data sample outputted by the two sub models to be consistent with each other, through minimizing the bidirectional Kullback-Leibler (KL) divergence \cite{bib32} between the two distributions. In this way, the inconsistency between the training and inference stage can be alleviated.

\subsection{Bidirectional Training}

Many studies have shown that pre-training can transfer the knowledge and data distribution, hence improving the generalization. Bidirectional training (BiT) \cite{bib10} happens to be a simple and effective pre-training method to improve the translation quality of NMT models. Bidirectional training is divided into two stages, the early stage bidirectionally updates model parameters, and then tune the model normally. To achieve bidirectional updating, we only need to reconstruct the training samples from "src$\rightarrow$tgt" to "src+tgt$\rightarrow$tgt+src" without any complicated model modifications. Notably, BiT does not increase any parameters or training steps, requiring the parallel data merely.

\subsection{Data Diversification}

Data Diversification (DD) \cite{bib5} is a data augmentation method to boost NMT performance. It diversifies the training data by using the predictions of multiple forward and backward models and then merging them with the original dataset on which the final NMT model is trained. DD is applicable to all NMT models. It does not require extra monolingual data, nor does it add more computations and parameters. To conserve training resources, we only use one forward model and one backward model when using DD.

\subsection{Forward Translation}

Forward translation (FT), also known as self-training \cite{bib6}, is one of the most commonly used data augmentation methods. FT has proven effective for improving NMT performance by augmenting model training with synthetic and authentic parallel data. Generally, FT is performed in three steps: (1) randomly sample a subset from the large-scale source monolingual data; (2) use a “teacher” NMT model to translate the subset data into the target language to construct the synthetic parallel data; (3) combine the synthetic and authentic parallel data to train a “student” NMT model.

\subsection{Back-Translation}

An effective method to improve NMT with target monolingual data is to augment the parallel training data with back translation (BT) \cite{bib7,bib40}. There are many works broaden the understanding of BT and investigates a number of methods to generate synthetic source sentences. Edunov et al. \cite{bib29} find that back translations obtained via sampling or noised beam outputs are more effective than back translations generated by beam or greedy search in most scenarios. Caswell et al. \cite{bib30} show that the main role of such noised beam outputs is not to diversify the source side, but simply to indicate to the model that the given source is synthetic. Therefore, they propose a simpler technique, Tagged BT. This method uses an extra token to mark back translated source sentences, which is generally outperform than noised BT.

\subsection{Alternated Training}

While synthetic bilingual data have demonstrated their effectiveness in NMT, adding more synthetic data often deteriorates translation performance since the synthetic data inevitably contains noise and erroneous translations. Alternated training (AT) \cite{bib8} introduce authentic data as guidance to prevent the training of NMT models from being disturbed by noisy synthetic data. AT describes the synthetic and authentic data as two types of different approximations for the distribution of infinite authentic data, and its basic idea is to alternate synthetic and authentic data iteratively during training until the model converges.

\subsection{Curriculum Learning}

A practical curriculum learning (CL) \cite{bib33} method should address two main questions: how to rank the training examples, and how to modify the sampling procedure based on this ranking. For ranking, we choose to estimate the difficulty of training samples according to their domain feature \cite{bib11}. The calculation formula of domain feature is as follows, where $\theta_{in}$ represents an in-domain NMT model, and $\theta_{out}$ represents a out-of-domain NMT model.

\begin{equation}
q(x,y)=\frac{\log{P(y\vert x;\theta_{in})}-\log{P(y\vert x;\theta_{out})}}{\vert y\vert} \label{eq1}
\end{equation}

For the sampling procedure, we adopt a probabilistic CL strategy\footnote{\url{https://github.com/kevinduh/sockeye-recipes/tree/master/egs/curriculum}} that takes advantage of the spirit of CL in a nondeterministic fashion without discarding the good practice of original standard training policy.

\subsection{Transductive Ensemble Learning}

Ensemble learning \cite{bib34}, which aggregates multiple diverse models for inference, is a common practice to improve the accuracy of machine learning tasks. However, it has been observed that the conventional ensemble methods only bring marginal improvement for NMT when individual models are strong or there are a large number of individual models. Transductive Ensemble Learning (TEL) \cite{bib12} study how to effectively aggregate multiple NMT models under the transductive setting where the source sentences of the test set are known. TEL uses all individual models to translate the source test set into the target language space and then finetune a strong model on the translated synthetic data, which boosts strong individual models with significant improvement and benefits a lot from more individual models.

\section{APE System}\label{sec_4}

\subsection{System Overview}

\begin{figure}[H] 
\centering
\includegraphics[width=100mm]{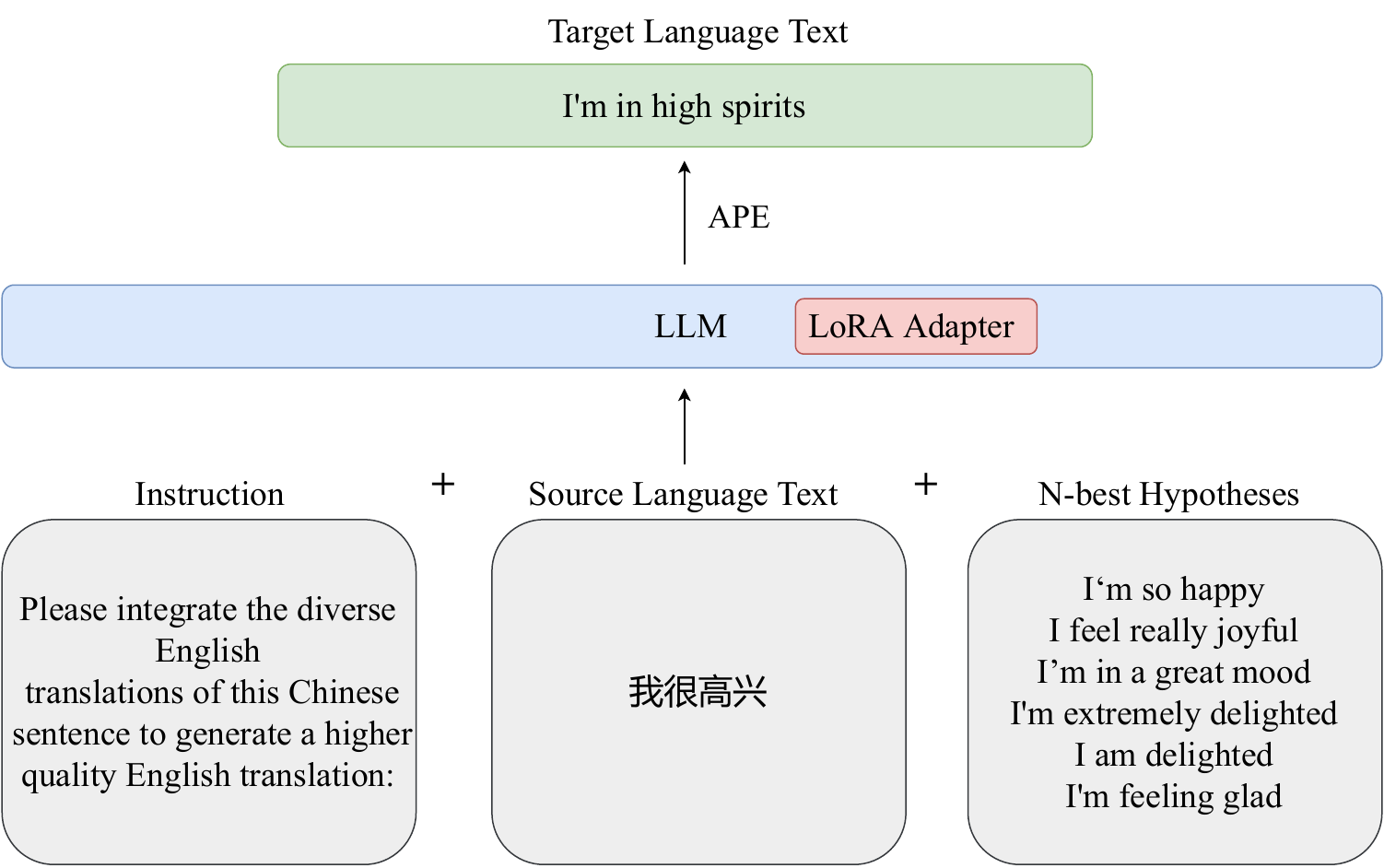}
\caption{\centering APE System for the zh$\rightarrow$en multi-domain machine translation task.}
\label{APE}
\end{figure}

There is recently a surge in research interests in Transformer-based LLMs, such as ChatGPT \cite{bib15,bib16} and LLaMA \cite{bib17,bib18,bib39}. Benefiting from the giant model size and oceans of training data, LLMs can understand better the language structures and semantic meanings behind raw text, thereby showing excellent performance in a wide range of natural language processing (NLP) tasks. As shown in Figure \ref{APE}, we use supervised fine-tuning to train LLM as an APE model to improve the translation quality of our NMT model on the zh$\rightarrow$en multi-domain machine translation task. Our APE system is inspired by the GenTranslate \cite{bib14}, but the difference is that we use source language text as part of the input information of LLM because we believe that adding source language text helps ensure the fidelity of the target language translation.

\subsection{Efficient LLM Finetuning}

We choose Llama2-13b as the base LLM for our APE system. When performing supervised fine-tuning on this base LLM, full fine-tuning by retraining all model parameters is usually expensive and requires a long training period. Therefore, we adopt the popular LoRA-based efficient parameter fine-tuning method\footnote{\url{https://github.com/microsoft/LoRA}} \cite{bib35}. This method freezes the pre-trained model weights and injects trainable rank decomposition matrices into each layer of the Transformer architecture, greatly reducing the number of trainable parameters for downstream tasks. LoRA can lower the hardware threshold by up to 3x when using an adaptive optimizer because we do not need to compute gradients or maintain optimizer state for most parameters.

\subsection{HypoTranslate Dataset}

To build the APE training data for Efficient LLM fine-tuning, we first use the cometkiwi model\footnote{\url{https://huggingface.co/Unbabel/wmt22-cometkiwi-da}} \cite{bib36} to select high-quality bilingual data. Specifically, we select bilingual data with a cometkiwi score greater than 0.8 on the zh$\rightarrow$en language pair. Then we use our trained NMT model as a base translation model to decode the N-best hypotheses from the source language text via a beam search algorithm, where the beam size N is set to 10.

\section{Experiments}\label{sec5}
\subsection{Setup}

We use Pytorch-based Fairseq \cite{bib37} open-source framework to train the NMT model, and use Adam optimizer with $\beta1$=0.9 and $\beta2$=0.98 to guide the parameter optimization. During the training phase, each model uses 8 GPUs for training, batch size is 2048, update frequency is 4, learning rate is 5e-4, label smoothing rate is 0.1 and warm-up steps is 4000. We set dropout to 0.1 for high-resource translation tasks and 0.3 for low-resource translation tasks respectively. In addition, when applying R-Drop for training, we follow the setting of L et al \cite{bib9}, using {\itshape reg\_label\_smoothed\_cross\_entropy} as the loss function, and set reg-alpha to 5. Then, we use SacreBLEU \cite{bib38} and COMET\footnote{\url{https://huggingface.co/Unbabel/wmt20-comet-da}} \cite{bib35} to evaluate the overall translation quality of each NMT model. 

To adapt LoRA-based efficient LLM fine-tuning \cite{bib35}, we use llama-recipes\footnote{\url{https://github.com/meta-llama/llama-recipes}} open source framework to train the APE model. The following is the configuration of LoRA: lora\_rank is 32, lora\_alpha is 64, lora\_dropout is 0.05, lora\_modules are {\itshape "q", "k", "v", "o", "gate", "down", "up"}. We train 2 epochs with AdamW optimizer \cite{bib36}, with learning rate initialized to 1e-4 and then linearly decrease to 1e-5 during training. The batch size is set to 6, with accumulation iterations set to 8 (i.e., real batch size is 48), the context length is 4096, and the batch strategy is packing.

\subsection{Bilingual MT Results}

\subsubsection{en$\rightarrow$zh \& zh$\rightarrow$en}

On en$\rightarrow$zh and zh$\rightarrow$en translation tasks, we use BiT and R-Drop to build a strong baseline system. Subsequently, we adopt the data augmentation methods of DD, FT and ST to improve the translation quality of baseline System. Next, we use AT guide model training with authentic bilingual data. Then, we use CL for domain adaptation. Finally, we train multiple NMT systems and integrate them using TEL as the final translation system. 

Table \ref{ECCE} shows the evaluation results of en$\rightarrow$zh and zh$\rightarrow$en translation systems. Compared with the baseline system, the final en$\rightarrow$zh and zh$\rightarrow$en translation systems improves significantly on CCMT 2022 test sets.

\begin{table}[ht]
\begin{center}
\caption{\centering BLEU and COMET scores of en$\rightarrow$zh \& zh$\rightarrow$en NMT system}\label{ECCE}%
\begin{tabular}{|l|cccc|c|c|c|c|}
\hline
 &&&& \multicolumn{2}{c|}{en$\rightarrow$zh}  & \multicolumn{2}{c|}{zh$\rightarrow$en}\\
\hline
CCMT 2022 test set &&&& BLEU & COMET & BLEU & COMET \\
\hline
BiT R-Drop baseline &&&& 54.37 & 0.6953  & 43.46 & 0.6754 \\
+ DD, FT \& ST &&&& 56.71 & 0.7203 & 44.57 & 0.6895 \\
+ AT &&&& 57.03 & 0.7358 & 44.74 & 0.6915 \\
+ CL &&&& 57.49 & 0.7527  & 46.83 & 0.7060 \\
+ TEL &&&& \textbf{57.71} & \textbf{0.7610} & \textbf{47.45} & \textbf{0.7264} \\
\hline
\end{tabular}
\end{center}
\end{table}

\subsubsection{mn$\rightarrow$zh, ti$\rightarrow$zh \& uy$\rightarrow$zh}

On mn$\rightarrow$zh, ti$\rightarrow$zh and uy$\rightarrow$zh translation tasks, we also use BiT and R-Drop to build a strong baseline system. The subsequent training method is similar to en$\rightarrow$zh and zh$\rightarrow$en translation tasks. The only difference is that we adopt DD and Tagged BT in the data augmentation stage, which is due to the lack of source language monolingual for these three tasks. Table \ref{MTU} is the evaluation results of mn$\rightarrow$zh, ti$\rightarrow$zh and uy$\rightarrow$zh translation systems on CCMT 2022 test set or CCMT 2023 test set. Overall, the final translation systems for all three tasks improves significantly compared to the baseline.

\begin{table}[ht]
\begin{center}
\caption{\centering BLEU scores of mn$\rightarrow$zh, ti$\rightarrow$zh and uy$\rightarrow$zh translation system}\label{MTU}%
\begin{tabular}{|l|ccccc|ccc|cccc|}
\hline
CCMT test set &&&&& mn$\rightarrow$zh (2023) &&& ti$\rightarrow$zh (2022) &&& uy$\rightarrow$zh (2023) &\\
\hline
BiT R-Drop baseline &&&&& 55.87 &&& 32.69 &&& 42.58&\\
+ DD \& Tagged BT &&&&& 59.73 &&& 36.52 &&& 49.31&\\
+ AT &&&&& 62.01 &&& 37.45 &&& 49.91&\\
+ CL &&&&& 62.96 &&& 38.40 &&& 50.54&\\
+ TEL &&&&& \textbf{63.59} &&& \textbf{40.90} &&& \textbf{51.27} &\\
\hline
\end{tabular}
\end{center}
\end{table}

\subsection{Multi-domain MT Results}

On the zh$\rightarrow$en multi-domain machine translation task, we first select the best model (TEL) in the zh$\rightarrow$en bilingual machine translation task as the baseline model. Then, we use the collected high-quality domain-related bilingual data to fine-tune for domain adaptation. Finally, we use the APE model based on llama-13b to improve the translation results generated by the NMT model, and use the post-editing result as the final translation result. 

Table \ref{bleu} and Table \ref{comet} shows the evaluation results of zh$\rightarrow$en multi-domain translation system on the dev set. It shows that domain adaptation is critical to multi-domain translation, and the powerful generation capability of LLM can further help improve the translation quality of the NMT model in the multi-domain translation task.

\begin{table}[ht]
\begin{center}
\caption{\centering BLEU scores of zh$\rightarrow$en multi-domain translation system}\label{bleu}
\begin{tabular}{|l|c|c|c|c|c|c|c|c|c|}
\hline
dev set & IT & car & electronic & energy & finance & literatrue & machine & medical & average\\
\hline
TEL & 30.72  & 30.08  & 42.57  & 32.94  & 33.35  & 36.78  & 37.38  & 49.51 & 36.67 \\
+ fine-tuning & 39.08 & 36.50 & 50.12 & 39.42 & 44.18 & 43.71 & 45.11 & 57.26 & 44.42 \\
+ APE & 43.3 & 39.59 & 50.76 & 39.59 & 47.75 & 47.71 & 50.09 & 59.82 & 47.33 \\
\hline
\end{tabular}
\end{center}
\end{table}

\begin{table}[ht]
\begin{center}
\caption{\centering COMET scores of zh$\rightarrow$en multi-domain translation system}\label{comet}
\begin{tabular}{|l|c|c|c|c|c|c|c|c|c|}
\hline
dev set & IT & car & electronic & energy & finance & literatrue & machine & medical & average\\
\hline
TEL & 0.4218  & 0.5406  & 0.6828  & 0.4785  & 0.5919  & 0.5128  & 0.5732  & 0.7747  & 0.5720 \\
+ fine-tuning & 0.5299 & 0.5788 & 0.7417 & 0.5850 & 0.6432 & 0.6689 & 0.6935 & 0.8227 & 0.6580 \\
+ APE & 0.5649 & 0.5868 & 0.7547 & 0.5857 & 0.6788 & 0.7012 & 0.7608 & 0.8221 & 0.6819 \\
\hline
\end{tabular}
\end{center}
\end{table}

\section{Conclusion}\label{sec6}

This paper presents HW-TSC's submission to the bilingual machine translation task and multi-domain machine translation task of CCMT 2024. For both translation tasks, we use a series of training strategies to train NMT models based on the deep Transformer-big architecture. Additionally, for the multi-domain machine translation task, we use the powerful generation capabilities of LLM to post-edit the translation results of the NMT model to obtain better translations. Relevant experimental results also show the effectiveness of our adopted strategies. By using these enhancement strategies, our submission achieves a competitive result in the final evaluation.

\end{document}